\documentclass[10pt,twocolumn,letterpaper]{article}

\usepackage{iccv}
\usepackage{times}
\usepackage{epsfig}
\usepackage{graphicx}
\usepackage{amsmath}
\usepackage{amssymb}

\usepackage{times}
\usepackage{microtype}
\usepackage{booktabs} 
\usepackage{caption}
\usepackage{subcaption}

\usepackage{lipsum}
\usepackage{epstopdf}
\usepackage{mathrsfs}
\usepackage{amsthm}
\usepackage{multirow}
\usepackage{bm}

\newtheorem{Theorem}{Theorem}
 
\newtheorem{Remark}{Remark} 
 
\usepackage{amsmath,amsfonts,bm}

\def\eqref#1{equation~\ref{#1}}

\def\1{\bm{1}}

\def\mA{{\bm{A}}}

\def\mC{{\bm{C}}}
\def\mD{{\bm{D}}}

\def\mI{{\bm{I}}}

\def\mL{{\bm{L}}}
\def\mM{{\bm{M}}}

\def\mU{{\bm{U}}}

\def\mX{{\bm{X}}}
\def\mY{{\bm{Y}}}
\def\mZ{{\bm{Z}}}

\DeclareMathAlphabet{\mathsfit}{\encodingdefault}{\sfdefault}{m}{sl}
\SetMathAlphabet{\mathsfit}{bold}{\encodingdefault}{\sfdefault}{bx}{n}
\newcommand{\tens}[1]{\bm{\mathsfit{#1}}}

\def\tX{{\tens{X}}}
\def\tY{{\tens{Y}}}

\usepackage[colorlinks,linkcolor=red, breaklinks=true,bookmarks=false]{hyperref}

\iccvfinalcopy 

\def\iccvPaperID{5528} 

\ificcvfinal\fi

\begin{document}

\title{Unifying Nonlocal Blocks for Neural Networks}
%

\author{Lei Zhu$^{1,3,4}$ 
~ Qi She$^{*, 2}$ 
~ Duo Li$^{7}$ 
~ Yanye Lu$^{1,3}$ 
~ Xuejing Kang$^{5}$ 
~ Jie Hu$^{6}$ 
~ Changhu Wang$^{2}$ \\
{\tt\small zhulei@stu.pku.edu.cn, \tt\small sheqi1991@gmail.com}
\and $^{1}$Institute of Medical Technology, Peking University Health Science Center, Peking University 
\and $^{2}$Bytedance AI Lab 
\and $^{3}$Department of Biomedical Engineering, Peking University 
\and $^{4}$Institute of Biomedical Engineering, Peking University Shenzhen Graduate School 
\and $^{5}$Beijing University of Posts and Telecommunications 
\and $^{6}$SKLCS \& University of Chinese Academy of Sciences 
\and $^{7}$Hong Kong University of Science and Technology
}

\maketitle
\ificcvfinal\thispagestyle{empty}\fi

\begin{abstract}
The nonlocal-based blocks are designed for capturing long-range spatial-temporal dependencies in computer vision tasks. Although having shown excellent performance, they still lack the mechanism to encode the rich, structured information among elements in an image or video.
In this paper, to theoretically analyze the property of these nonlocal-based blocks, we provide a new perspective to interpret them, where we view them as a set of graph filters generated on a fully-connected graph. Specifically, when choosing the Chebyshev graph filter, a unified formulation can be derived for explaining and analyzing the existing nonlocal-based blocks (e.g., nonlocal block, nonlocal stage, double attention block). Furthermore, by concerning the property of spectral, we propose an efficient and robust spectral nonlocal block, which can be more robust and flexible to catch long-range dependencies when inserted into deep neural networks than the existing nonlocal blocks. Experimental results demonstrate the clear-cut improvements and practical applicabilities of our method on image classification, action recognition, semantic segmentation, and person re-identification tasks. Code are available at \url{https://github.com/zh460045050/SNL_ICCV2021}.
\end{abstract}

\section{Introduction}

Capturing the long-range spatial-temporal dependencies between spatial pixels or temporal frames plays a crucial role in computer vision tasks. Convolutional neural networks (CNNs) are inherently limited by their convolution operators, which are devoted to concern local relations (\eg, a $7 \times 7$ region), and they are inefficient in modeling long-range dependencies. Deep CNNs model these dependencies, which commonly refers to enlarge receptive fields, via stacking multiple convolution operators. However, two unfavorable issues are raised in practice. Firstly, repeating convolutional operations comes with higher computational and memory costs as well as the risk of over-fitting~\cite{cost}. Secondly, stacking more layers cannot always increase the effective receptive fields~\cite{widelimit}, which indicates the convolutional layers may still lack the mechanism to model these dependencies efficiently.

A common practice to tackle these challenges is to aggregate the feature in a non-local way with fewer learning weights. Thus the aggregation can act on not only the k-hop neighbors but also the long-range positions~\cite{deeplab, nl, defnet, defnetv2, pspnet}. Typically, inspired by the self-attention strategy, the Nonlocal (NL) block~\cite{nl} firstly creates a dense affinity matrix that contains the relation among every pairwise position, and then uses this matrix as an attention map to aggregate the features by weighted mean. Nonetheless, because the dense attention map concerns humongous amounts of feature pairs (\eg the relations between background and background), the aggregation feature map contains too much noise.

To solve this problem, recent state-of-the-art methods focus on creating a more reasonable attention map for the NL block~\cite{a2net, ns, cgnl, ccnet, ns}. Chen \etal~\cite{a2net} propose the Double Attention (A$^2$) block that firstly gathers the features in the entire space and then distributes them back to each location.
Yue \etal~\cite{cgnl} propose the Compact Generalized Nonlocal (CGNL) block to catch cross-channel clues, which also increases the noise of the attention map inevitably. 
Huang \etal~\cite{ccnet} propose a lightweight nonlocal block called Criss-Cross Attention block (CC), which decomposes the position-wise attention of NL into conterminously column-wise and row-wise attention.
To enhance the stability of the NL block, Tao \etal~\cite{ns} propose the Nonlocal Stage (NS) module that can follow the diffusion nature by using the Laplacian of the affinity matrix as the attention map. 
\begin{figure*}[!htbp]
    \centering
    \includegraphics[width=0.88\textwidth]{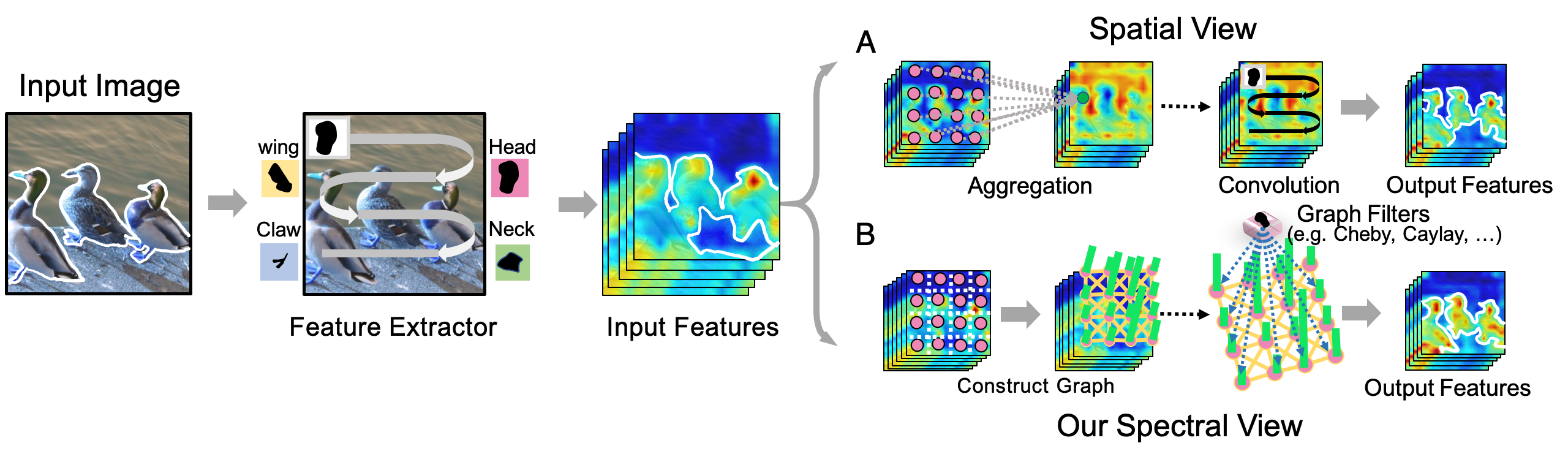}
    \caption{The spatial (\textbf{\sf{A}}) and spectral (\textbf{\sf{B}}) view of a nonlocal block. The pink dots indicate each patch in the feature map and the ``Aggregation" means calculating the weighted mean as the numerator of Eq.~(\ref{org_nlop}). The dotted arrows mean ``copy" and full arrows mean ``feed forward". The green bars are the node features and the length means their strength (best view in color).}
    \label{intro}
\end{figure*}

In general, though the above works can generate a better aggregation feature map by improving the attention mechanism of the nonlocal block, their parameterized filter step still lacks the long-range dependence concerning where only localization filters (usually $1\times1$ convolutions) are used as shown in Fig.~\ref{intro} \textcolor{red}{\textbf{\sf{A}}}. To conquer the limitations, we aim at combining the nonlocal aggregation with the local filter step of nonlocal block via graph filters. We call this perspective ``the spectral view of the nonlocal block". Specifically, as shown in Fig.~\ref{intro} \textcolor{red}{\textbf{\sf{B}}}, a fully-connected graph is firstly constructed based on the input features to contain the similarity among different positions. Then the output is generated by directly filtering the features with learnable graph filters. Under this perspective, long-range cues are maintained by the affinity of graph structure, and the graph filters provide more learning flexibility when catching long-range dependencies.

Moreover, the proposed ``spectral view" also provides a valid tool to theoretically analyze the nonlocal-based blocks~\cite{nl,ns,a2net,ccnet,cgnl}, which are all experimentally designed and lacks theoretical supports to explain their effectiveness. With the proposed ``spectral view", a unified formulation of these nonlocal-based blocks can be deduced. It shows that they all utilize ``an incomplete formation'' Chebyshev graph filter, which limits their performance and robustness. Profited by our ``spectral view", these incomplete aspects can be easily solved by concerning the graph nature. Hence, a novel nonlocal block called the Spectral Nonlocal Block (SNL) with a stronger theoretical basis is proposed, which uses a more accurate approximation and can avoid complex eigenvalue. The proposed SNL is more flexible when inserted into the neural network and achieves superior performance over other nonlocal-based blocks. This shows the effectiveness of the proposed ``spectral view" in defining novel nonlocal blocks on the basis of graph signal processing~\cite{gsp} or with the help of graph neural networks~\cite{gcn, cay_net, survey_gcn}.

In a nutshell, our contributions are threefold: 
\begin{itemize}
\item {We provide a novel perspective for the model design of nonlocal-based blocks, which can help to catch long-range dependencies more flexibly. In this context, we seamlessly make the connection between different variants of nonlocal methods and graph signal processing.}
\item {We present, for the first time, that five well-known nonlocal blocks can be unified and interpreted under the proposed perspective with the help of Chebyshev approximation. This motivates our advocate of other potential nonlocal blocks embodied with discrimination capability and theoretical basis as an alternative.}
\item {We propose a novel nonlocal block with a stronger theoretical basis that using an accurate approximated filter with the concern of the graph nature. The proposed block works universally well across a wide array of vision tasks, including image classification, action recognition, semantic segmentation, and person Re-identification, offering significantly better performance than other nonlocal-based counterparts.}
\end{itemize}

\section{Preliminary}
In this paper, \textbf{bold} we use uppercase characters to denote the parameter matrix and \textit{\textbf{italic bold}} uppercase characters to denote other matrices. Vectors are denoted with lowercase. 

\subsection{Nonlocal Block} 
The NL block calculates a weighted sum of pairings between the features of each position and all possible positions as shown in Fig.~\ref{intro} \textcolor{red}{\textbf{\sf{A}}}. The nonlocal operator is defined:
\begin{equation}
    F(\mX_{i, :}) = \frac{\sum_{j}\Big[f(\mX_{i, :}, \mX_{j, :})g(\mX_{j, :})\Big]}{\sum_{j} f(\mX_{i, :}, \mX_{j, :})}, 
\label{org_nlop}
\end{equation}
where $\mX \in \mathbb{R}^{N \times C_{1}}$ is the input feature map, $i, j$ are the position indexes in the feature map, $f(\cdot)$ is the affinity kernel with a finite Frobenius norm. $g(\cdot)$ is a linear embedding that is defined as: $g(\mX_{j, :}) = \mX_{j, :} \mathbf{W}_{Z}$ with $\mathbf{W}_{Z} \in \mathbb{R}^{C_{1} \times C_{s}}$. Here $N$ is the total positions of each feature. $C_1$ and $C_s$ are the numbers of channels for the input and the transferred features, respectively. 

When inserting the NL block into the network structure, a linear transformation with weight matrix $\mathbf{W} \in \mathbb{R}^{C_{s} \times C_{1}}$ and a residual connection are added:
\begin{equation}
    \mY_{i, :} = \mX_{i, :} + F(\mX_{i, :}) \mathbf{W}.
\label{org_nl}
\end{equation}

It is worth noting that, though NL block and the Vision Transformers (ViT)~\cite{VIT, SWIN, VITSURV} both utilize the self-attention mechanism, the former is added on a certain stage of CNNs to perceive long-range dependencies rather than using to replace all convolutional operators in CNN as the latter.

\subsection{Graph Fourier Transform \& Graph Filter}

Assuming that a graph $\mathcal{G}$ contains $N$ vertices, an arbitrary function (or signal) vector $\bm{f} = \{f_1, f_2, \cdots, f_N\}$ can be defined, where the $i^{th}$ component of the vector $\bm{f}(i)$ represents the function value at the $i^{th}$ vertex of the graph $\mathcal{G}$. Then, the graph Fourier  transformation~\cite{gsp} $\hat{\bm{f}}$ of $\bm{f}$ on vertices of $\mathcal{G}$ can be formulated as the expansion of $\bm{f}$ in terms of the eigenvectors of the graph Laplacian:
\begin{equation}
\hat{\bm{f}}(\lambda_{l}) = \sum^N_{i=1} \bm{f}(i) \bm{u}^{*}_l(i), 
\end{equation}
where $\bm{\lambda}=[\lambda_1, \cdots, \lambda_l, \cdots]$ and $\mU=[\bm{u}_1, \cdots \bm{u}_l, \cdots]$ are the eigenvalues and eigenvectors of the graph Laplacian. $\hat{\bm{f}}(\lambda_{l})$ is the corresponding spectral coefficient under $\bm{u}_l$. $\bm{u}^{*}_l$ is the $l^{th}$ row vector of $\mU^{\top}$. The inverse graph Fourier transformation is then given by $\bm{f}(i) = \sum^N_{i=1} \hat{\bm{f}}(\lambda_l) \bm{u}_l(i)$.  

A graph filter is an operator that modifies the components of an input signal $\bm{x}$ based on the eigenvalues $\mU$, according to a frequency response $\mathbf{g}_{\theta}$ acting on  $\bm{\lambda}$. Thus, based on the graph Fourier transformation, the output of filtering $\bm{x}$ under graph $\mathcal{G}$ can be defined as:
\begin{equation}
\bm{O}(i)_{\bm{x} *_{\mathcal{G}} \mathbf{g}_{\theta}} = \sum^N_{l=1} \hat{\bm{O}}(\lambda_l) \bm{u}_l(i) = \sum^N_{l=1} \hat{\bm{x}}(\lambda_l) \hat{\mathbf{g}}_{\theta}(\lambda_l) \bm{u}_l(i)
\end{equation}
\noindent where $\hat{\bm{x}}$, $\hat{\mathbf{g}}_{\theta}$, $\hat{\bm{O}}$ are the graph Fourier transformation of input signal $\bm{x}$, filter $\mathbf{g}_{\theta}$ and the output signal $\bm{O}_{\bm{x} *_{\mathcal{G}} \mathbf{g}_{\theta}}$ respectively. Further, the formulation of the output signal can be also derived as (more details of this derivation and graph signal process refer to the survey~\cite{gsp}):
\begin{equation}
\bm{O}_{\bm{x} *_{\mathcal{G}} \mathbf{g}_{\theta}} = \mU \mathrm{diag}([\hat{\mathbf{g}}_{\theta}(\lambda_1), \cdots, \hat{\mathbf{g}}_{\theta}(\lambda_l), \cdots]) \mU^{\top} \bm{x}
 \end{equation}

\section{Approach}
\begin{table*}[!htbp]
\centering
\small
\caption{Summary of five existing nonlocal-based blocks in the spectral view.}
\begin{tabular}{c|c|c|c|c|c}
\toprule
Models & Vertex ($|\mathbb{V}|$)& Edge ($|E|$)  & Affinity Matrix ($\mA$) & Node Feature ($\mZ$)& Formulation of ${F}(\mA, \mZ)$\\
\hline
\textbf{Chebyshev}  & - & - & - & - & $\mZ \mathbf{W}_{1} + \mA \mZ \mathbf{W}_{2} + \sum_{k=2}^{K-1} \mA^k \mZ \mathbf{W}_{k}$\\
\hline
NL & $N$ & $N \times N$ & $\mD^{-1}_{M} \mM$ & $\mX \mathbf{W}_Z$ & $\mA \mZ \mathbf{W}$ \\
\hline
$\mathrm{A^2}$ & $N$ & $N \times N$ & $\bm{M}$ & $\mX \mathbf{W}_Z$ & $\mA \mZ \mathbf{W}$ \\
\hline
CGNL & $NC_s$ & $NC_s \times NC_s$ & $\mD^{-1}_{M^{f}} \mM^{f}$ & $\mathrm{vec}(\mX \mathbf{W}_Z)$ & $\mA \mZ \mathbf{W}$ \\
\hline
NS & $N$ & $N \times N$ &  $\mD^{-1}_{M} \mM$ & $\mX \mathbf{W}_Z$ & $ -\mZ \mathbf{W} + \mA \mZ \mathbf{W}$ \\
\hline
CC & $N$ & $N \times N$ & $\mD^{-1}_{\mC \odot \mM} (\mC \odot \mM)$ & $\mX$ & $\mA \mZ \mathbf{W}$ \\
\bottomrule
\end{tabular}
\label{relation}
\end{table*}
The nonlocal operator can be explained under the graph spectral domain, where it is the same as operating a set of graph filters on a fully connected weighted graph. This process can be briefly divided into two steps: 1) generating a fully-connected graph to model the similarity among the position pairs, and 2) converting the input features into the graph domain and learning a graph filter. In this section, we firstly propose our framework which gives the definition of the spectral view for the nonlocal operator. Then, we unify five existing nonlocal-based operators from this spectral view. We further propose a novel nonlocal block based on the framework, which is more effective and robust.

\subsection{The Spectral View of Nonlocal-based blocks} 
\label{sec_3_1}
To define the spectral view of nonlocal, we start from taking the matrix form of the nonlocal operator into Eq.~(\ref{org_nl}) and decompose the parameter matrix $\mathbf{W}$ into $\mathbf{W}_{s1}$ and  $\mathbf{W}_{s2}$:
\begin{align}
\mY = \mX + F(\mX)\mathbf{W} = \mX + \mA \mZ \mathbf{W}_{s1}\mathbf{W}_{s2},
\label{nl_eq}
\end{align}
In this Eq.~(\ref{nl_eq}), $\mA=\mD^{-1}_{M} \mM$ is the affinity matrix and $\mM = [M_{ij}]$ is composed of pairwise similarities between pixels, \ie $M_{ij} = f(\mX_{i,:}, \mX_{j,:})$ where $f(\cdot)$ is usually the dot product. $\mD_{M}$ is a diagonal matrix containing the degree of each vertex of $\mM$. $\mZ= \mX \mathbf{W}_{Z} \in \mathbb{R}^{N \times C_{s}}$ is the transferred feature map that compresses the channels of $\mX$ by a linear transformation with $\mathbf{W}_Z \in \mathbb{R}^{C_{1} \times C_{s}}$. $\mathbf{W}_{s1} \in \mathbb{R}^{C_s \times C_s}$ and $\mathbf{W}_{s2} \in \mathbb{R}^{C_s \times C_1}$ are two parameter-matrices that are used to filter discriminative features and restore the number of channels respectively. Then based on Eq.~(\ref{nl_eq}), the nonlocal block can be formulated in the spectral view by generalizing $\mathbf{W}_{s1}$ into a set of graph filters $\mathbf{g}_{\theta} = \{\mathbf{g}^1_{\theta}, \cdots, \mathbf{g}^i_{\theta}, \cdots, \mathbf{g}^{C_s}_{\theta}\}$:
\begin{equation}
    \mY =  \mX + \mathcal{F}(\mA, \mZ, \mathbf{W}_{s1})\mathbf{W}_{s2} =  \mX + \mathcal{F}(\mA, \mZ, \mathbf{g}_{\theta})\mathbf{W}_{s2},
\label{eq:nl}
\end{equation}
where $\mathcal{F}(\mA, \mZ, \mathbf{g}_{\theta})$ is the ``nonlocal operator in the spectral view". For clarity, we omit $\mathbf{W}_{s2}$ by assuming $C_1 = C_s$, abbreviate $\mathcal{F}(\mA, \mZ, \mathbf{g}_{\theta})$ into $\mathcal{F}(\mA, \mZ)$ and call it ``nonlocal operator" in following paper.

In this view, the nonlocal operator firstly computes the affinity matrix $\bm{A}$ that defines a graph spectral domain and then learns filters $\mathbf{g}_{\theta}$ for graph spectral features. Specifically, a fully-connected graph $\mathcal{G}=\{\mathbb{V}, \bm{A}, \bm{Z}\}$ is constructed, in which $\mathbb{V}$ is the vertex set. Then, for each column vector $\bm{z}_{i} \in \mathbb{R}^{N\times1}$ of $\mZ$, a graph filter $\mathbf{g}^{i}_{\theta}$ is generated to enhance the feature discrimination. From this perspective, the nonlocal operator can be theoretically interpreted in the spectral view as below:
\begin{Theorem}
Given an affinity matrix $\mA \in \mathbb{R}^{N \times N}$ and the signal $\mZ \in \mathbb{R}^{N \times C}$, the nonlocal operator is the same as filtering the signal $\mZ$ by a set of graph filters $\{\mathbf{g}^{i}_{\theta}, \mathbf{g}^{2}_{\theta}, \cdots, \mathbf{g}^{C}_{\theta}\}$ under the graph domain of a fully-connected graph $\mathcal{G}$:
\begin{align}
        \mathcal{F}(\mA, \mZ) = [\bm{O}_{\bm{z}_{1} *_{\mathcal{G}} \mathbf{g}^1_{\theta}}, \cdots, \bm{O}_{\bm{z}_{i} *_{\mathcal{G}} \mathbf{g}^{i}_{\theta}}, \cdots, \bm{O}_{\bm{z}_{C} *_{\mathcal{G}} \mathbf{g}^{C}_{\theta}} ]
\end{align}
where the graph $\mathcal{G} = (\mathbb{V}, \mZ, \mA)$ has the vertex set $\mathbb{V}$, node feature $\mZ$ and affinity matrix $\mA$. $\bm{O}_{\bm{z}_{i} *_{\mathcal{G}} \mathbf{g}^{i}_{\theta}} \in \mathbb{R}^{N \times 1}$ is the output signal on $\bm{z}_{i}$.
\label{T.1}
\end{Theorem}

\begin{Remark}
The main difference between Theorem.~\ref{T.1} and the original spatial view of nonlocal~\cite{nl} is that the former learns a graph filter to obtain the feature under the spectral domain while the latter filters the feature by the convolutional operator without concerning the graph structure. Moreover, to confirm the existence of the graph spectral domain, Theorem.\ref{T.1} requires that the graph Laplacian $\mL$ should be diagonalized and not contain complex eigenvalues and eigenvectors. Thus the affinity matrix $\mA$ should be symmetric.
\end{Remark}

Specifically, a generalized implementation~\cite{cheb_net} of the output signal on each column vector $\bm{z}_{i}$ can be used for Theorem.~\ref{T.1} by setting the graph spectral filter as a set of diagonal parameter matrix $\mathbf{\Omega}^{i} \in \mathbb{R}^{N \times N}$:
\begin{equation}
     \bm{O}_{\bm{z}_{i} *_{\mathcal{G}} \mathbf{g}^{i}_{\theta}} = \mU \mathbf{\Omega}^{i}\mU^\top \bm{z}_{i}
\label{eq:graphview}
\end{equation}
\noindent where $\mathbf{\Omega}^{i} = \mathrm{diag}([\omega_1, \omega_2, \cdots, \omega_n])$ contains $n$ parameters. In addition, new nonlocal operators can also be theoretically designed by using different types of graph filters to obtain output signal in Theorem.~\ref{T.1}, such as Chebyshev filter~\cite{cheb_net, kipf2016semi}, graph wavelet filter~\cite{wavelets}, Cayley filter~\cite{cay_net}.

\subsection{Unifying Existing Nonlocal-based Blocks}
The proposed ``spectral view" provides a valid tool to analyze the experimental designed nonlocal block on basis of graph signal processing. To unify other nonlocal-based blocks based on Theorem.~\ref{T.1}, here we use the Chebyshev filter for illustration (Cayley filter is also presented in appendix 2), \ie using Chebyshev polynomials~\cite{cheb_net} to reduce the $n$ parameters in $\mathbf{\Omega}^{i}$ into $K$ ($K$ is the order of polynomials, and $K \ll N$). For simplicity, we firstly assume that both the input and output signals have one channel, \ie $\mZ = \bm{z}_{1}$ and $\mathcal{F}(\mA, \mZ) = \mU \mathbf{\Omega}^{1} \mU^\top \bm{Z}$. Then the parameter matrix $\mathbf{\Omega}^{1}$ of the graph filter approximated by $K^{th}$-order Chebyshev polynomials is formulated as:
\begin{align}
  \mathcal{F}(\mA, \mZ) &= \sum_{k=0}^{K-1} \hat{\theta}_k T_k(\bm{\widetilde{L}}) \mZ,\label{eq:snl_1}\\
  \mathrm{s.t.} \quad
T_k({\bm{\widetilde{L}}}) &= 2 \bm{\widetilde{L}} T_{k-1}(\bm{\widetilde{L}}) - T_{k-2}(\bm{\widetilde{L}}),\nonumber
\end{align}
where $\bm{\widetilde{L}} = 2\bm{L} / \lambda_{\mathrm{max}} - \mI_{n}$, $T_{0}(\bm{\widetilde{L}}) = \mI_{N}$, $T_1({\bm{\widetilde{L}}}) = \bm{\widetilde{L}}$, and $\hat{\theta}_{k}$ is the coefficient.

Note that the affinity matrix $\bm{A}$ is affected by the input feature $\mX$ rather than using a fixed graph structure. Thus, an upper-bound exists for the maximum eigenvalue on all possible affinity matrices, \ie $\lambda_{\mathrm{max}} = 2$, when all their graph Laplacian $\bm{L}$ are normalized graph Laplacian~\cite{gsp}. With this assumption, we can get $\bm{\widetilde{L}} = -\mA$ and take it into Eq.~(\ref{eq:snl_1}):
\begin{equation}
    \mathcal{F}(\mA, \mZ) = \theta_{0} \mZ + \theta_{1} \mA \mZ + \sum_{k=2}^{K-1} \theta_k \mA^{k} \mZ, 
\label{spectral_nonlocal_k}
\end{equation}
where $\theta_{k}$ can be learned via SGD. Then, extending Eq.~(\ref{spectral_nonlocal_k}) into multiple channels, we can get a generalized formulation of the nonlocal operator with Chebyshev filter:
\begin{equation}
\mathcal{F}(\mA, \mZ) = \mZ \mathbf{W}_{1} + \mA \mZ \mathbf{W}_{2} + \sum_{k=2}^{K-1} \mA^k \mZ \mathbf{W}_{k+1},
\label{full_SNL}
\end{equation}
where ${F}(\mA, \mZ)$ is the nonlocal operator, $\mathbf{W}_{k} \in \mathbb{R}^{C_{s} \times C_{s}}$ is a parameter matrix. Note that, when $C_{s} \neq C_{1}$, it is straightforward to merge $\mathbf{W}_{s2}$ with $\mathbf{W}_{k}$, which makes $\mathbf{W}_{k}= \mathbf{W}_{k}*\mathbf{W}_{s2} \in \mathbb{R}^{C_{s} \times C_{1}}$.

Eq.~(\ref{full_SNL}) gives the connection between spatial view and spectral view of the nonlocal operator, in which the graph filter is expressed by the aggregation among the $k^{th}$ neighbor nodes. Thus, existing nonlocal-based structures can be theoretically analyzed by Eq.~(\ref{full_SNL}) in the spectral view. Here we elaborate $5$ types of existing nonlocal-based blocks that can be unified. They can be interpreted under certain graph structures and assumptions as shown in Table~\ref{relation}. More derivation details can be found in appendix 1.

\noindent (1) NL block~\cite{nl}: The NL block in the spectral view is the same as defining the graph as $\mathcal{G}=(\mathbb{V}, \mD^{-1} \mM, \mZ)$ and then using the second term of the Chebyshev polynomials to approximate the generalized graph filter.


\noindent (2) NS module~\cite{ns}: The NS module in the spectral view can be considered as the graph in the form of $\mathcal{G}=(\mathbb{V}, \mD^{-1}_{M} \mM, \mZ)$. The $1^{st}$-order Chebyshev polynomials is utilized to approximate the graph filter with the condition $\mathbf{W}_1 = - \mathbf{W}_2 = - \mathbf{W}$.


\noindent (3) A$^2$ block~\cite{a2net}:
The Double Attention Block can be viewed as the graph $\mathcal{G}=(\mathbb{V}, \bm{M}, \mZ)$ and then we can use the second term of the Chebyshev polynomial to approximate the graph filter, i.e $F(\mA, \mZ) = \bm{M} \bm{Z} \mathbf{W}$.

\begin{figure*}[!htbp]
    \centering
    \includegraphics[width=0.86\textwidth]{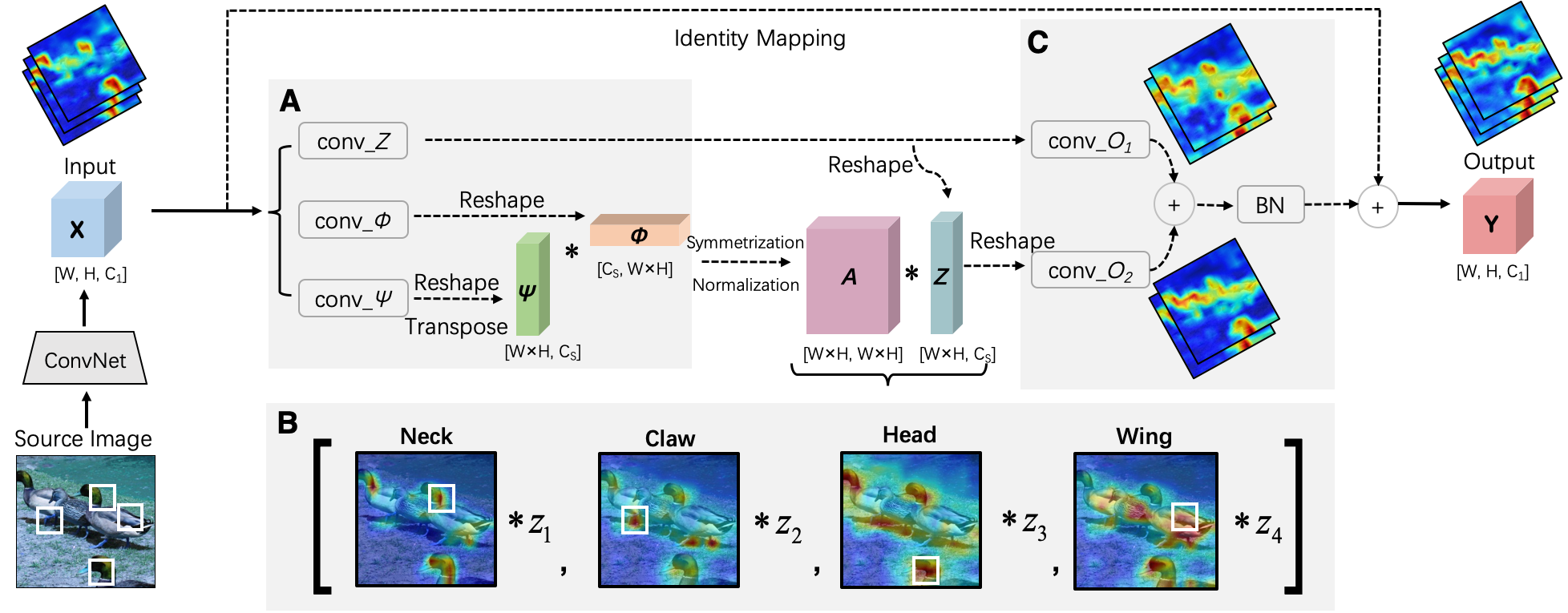}
    \caption{The implementation of our SNL. \textbf{ \sf{A}.} Three feature maps $\bm{\phi}$, $\bm{\psi}$, $\mZ$ are generated by feeding the $\tX$ into three $1\times1$ convolutions. \textbf{ \sf{B}.} The second term of Eq.~(\ref{snl}) is calculated with $\mZ$ and a normalized symmetrization affinity matrix $\mA$. Each row of $\mA$ contains a $N$-dimension spatial attention map (heat maps) and $\mathbf{z}_{1}, \mathbf{z}_{2}, \cdot\cdot\cdot, \mathbf{z}_{n}$ are the column vectors of $\mZ$ (for simplicity, here we pick $n=4$ where the white squares are the central positions we visualize). \textbf{\sf{C}.} The graph filter is approximated by respectively feeding the $0^{th}$-order and $1^{st}$-order term into convolutions to obtain the output $\tY$.}
    \label{gnl_block}
\end{figure*}


\noindent (4) CGNL block~\cite{cgnl}:
When grouping all channels into one group, the CGNL can be viewed as the graph $\mathcal{G}=(\mathbb{V}^f, \mD^{-1}_{M^f} \mM^f, \text{vec}(\mZ))$, where $\mathbb{V}^f$ contain both spatial position and feature channel. Again we can also use the second term of the Chebyshev Polynomial to approximate the graph filter, i.e $F(\mA, \mZ) =  \mD^{-1}_{M^f} \mM^f \text{vec}(\mZ) \mathbf{W}$.


\noindent (5) CC block~\cite{ccnet}:
In the CC block, $\mathcal{G}=(\mathbb{V}, \mD^{-1}_{\mC \odot \mM} \mC \odot \mM, \mX)$ with mask $\mC$ and then the second term of the Chebyshev is used to approximate filters with node feature $\mX$.

\subsection{Designing Spectral Nonlocal Blocks}
\label{sec_3_3}
Except for unifying the existing nonlocal blocks, our proposed ``spectral view" can also help to design novel nonlocal-based blocks with theoretical guarantees. As an example, we elaborate a more rational nonlocal block called Spectral Nonlocal Block (SNL), which is still based on the \textbf{Chebyshev filter} but has a stronger theoretical basis and robustness. 

Based on the above section, existing nonlocal-based operators use the random walk normalized (NL, NS, CGNL, CC) or the non-normalized affinity matrix ($\mathrm{A^2}$). The symmetry property is not guaranteed and the performance depends on the affinity kernel. This leads to the non-existence of the graph spectral domain and causes large amounts of complex eigenvalues for the graph Laplacian as discussed in Sec.\ref{sec_3_1}. This problem is also magnified under the fully-connected graph model of the nonlocal-based blocks.

Moreover, existing nonlocal blocks only use the second term (NL, $\mathrm{A^2}$, CGNL, CC) or the $1^{st}$-order approximation with sharing weight (NS) rather than the complete form of the $1^{st}$-order approximation, which also hinders their performance. Thus considering these factors, the proposed SNL block applies a symmetry affinity matrix with a more complete approximation as:
\begin{align}
    &\bm{Y} = \mX + \mathcal{F}_s(\mA, \mZ) = \mX + \mZ \mathbf{W}_{1} + \mA \mZ \mathbf{W}_{2}, \label{snl}\\
    &\mathrm{s.t.}  \quad \mA = \mD^{-\frac{1}{2}}_{\hat{M}} \bm{\hat{M}} \mD^{-\frac{1}{2}}_{\hat{M}} ,\quad \bm{\hat{M}} = (\mM + \mM^{\top}) / 2 \nonumber
\end{align}
where $\mathcal{F}_{s}(\mA, \mZ)$ is the SNL operator, $\mathbf{W}_{1}$, $\mathbf{W}_{2}$ are two parameter matrices.

\begin{Remark}
The proposed SNL uses Chebyshev filter as the generalized graph filter but has a more thorough formulation analyzed by our proposed spectral view. Specifically, it uses a symmetric affinity matrix $\mA = \mD^{-\frac{1}{2}}_{\hat{M}} \bm{\hat{M}} \mD^{-\frac{1}{2}}_{\hat{M}}$ to ensure the existence of the real eigenvalue. This makes it more stable when inserted into the deep neural networks. Moreover, the proposed SNL also uses the complete form of $1^{st}$-order Chebyshev Approximation that is a more accurate approximation of the graph filter. Thus, it can give the parameters a liberal learning space with only one more parameter matrix.
\end{Remark}

The implementation details of the SNL block are shown in Fig.~\ref{gnl_block} $(\textcolor{red}{\textbf{\sf{A}}, \textbf{\sf{B}}, \textbf{\sf{C}}})$. The input feature map $\tX \in \mathbb{R}^{W \times H \times C_1}$ is firstly fed into three 1$\times$1 convolutions with the weight kernels: $\mathbf{W}_{\phi,\psi,g} \in \mathbb{R}^{C_{1} \times C_{s}}$ to subtract the number of channels and then be reshaped into $\mathbb{R}^{WH \times C_{s}}$. One of the output $\mZ \in \mathbb{R}^{WH \times C_{s}}$ is used as the transferred feature map to reduce the calculation complexity, while the other two outputs $\bm{\varPhi}, \bm{\varPsi} \in \mathbb{R}^{WH \times C_{s}}$ are used to get the affinity matrix $\mA$ with the affinity kernel function $f(\cdot)$. Then, $\mA$ is made to be symmetric and normalized as in Eq.~(\ref{snl}). Finally, with the affinity matrix $\mA$ and the transferred feature map $\mZ$, the output of the nonlocal block can be obtained by the Eq.~(\ref{snl}). Specifically, the two weight matrices $\mathbf{W}_{1,2} \in \mathbb{R}^{C_{s} \times C_{1}}$ are yielded by two 1$\times$1 convolutions.

\begin{table*}[!htbp]
\centering
\small
\caption{The Performance of Nonlocal-based Blocks with Different Number of Transferred Channels on CIFAR-100}
\label{Tab:subchannel}
\renewcommand\tabcolsep{3.0pt}
\begin{tabular}{lcc|lcc|lcc}
\toprule
\multicolumn{3}{c|}{Non-Reduction} & \multicolumn{3}{c|}{Half-Reduction} & \multicolumn{3}{c}{Quarter-Reduction}\\
\hline
Models & Top-1 (\%) & Top-5 (\%) & Models & Top-1 (\%) & Top-5 (\%)  & Models & Top-1 (\%) & Top-5 (\%) \\
\hline
PreResNet56 & $ 75.33^{\textcolor{red}{\uparrow}0.00}$ & $ 93.97^{\textcolor{red}{\uparrow}0.00}$ & 
PreResNet56 & $ 75.33^{\textcolor{red}{\uparrow}0.00}$ & $ 93.97^{\textcolor{red}{\uparrow}0.00}$ &
PreResNet56 & $ 75.33^{\textcolor{red}{\uparrow}0.00}$ & $ 93.97^{\textcolor{red}{\uparrow}0.00}$ \\

+ NL & $ 75.29^{\textcolor{green}{\downarrow}0.04}$ & $ 94.07^{\textcolor{red}{\uparrow}0.10}$ & 
+ NL & $ 75.31^{\textcolor{green}{\downarrow}0.02}$ & $ 92.84^{\textcolor{green}{\downarrow}1.13}$ &
+ NL & $ 75.50^{\textcolor{red}{\uparrow}0.17}$ & $ 93.75^{\textcolor{green}{\downarrow}0.22}$ \\

+ NS & $ 75.39^{\textcolor{red}{\uparrow}0.06}$ & $ 93.00^{\textcolor{green}{\downarrow}0.97}$ & 
+ NS & $ 75.83^{\textcolor{red}{\uparrow}0.50}$ & $ 93.87^{\textcolor{green}{\downarrow}0.10}$ &
+ NS & $ 75.61^{\textcolor{red}{\uparrow}0.28}$ & $ 93.66^{\textcolor{green}{\downarrow}0.31}$ \\

+ $\mathrm{A^2}$  & $ 75.51^{\textcolor{red}{\uparrow}0.18}$ & $ 92.90^{\textcolor{green}{\downarrow}1.07}$ & 
+ $\mathrm{A^2}$  & $ 75.58^{\textcolor{red}{\uparrow}0.25}$ & $ 94.27^{\textcolor{red}{\uparrow}0.30}$ &
+ $\mathrm{A^2}$   & $ 75.61^{\textcolor{red}{\uparrow}0.28}$ & $ 93.61^{\textcolor{green}{\downarrow}0.36}$ \\

+ CGNL & $ 74.71^{\textcolor{green}{\downarrow}0.62}$ & $ 93.60^{\textcolor{green}{\downarrow}0.37}$ & 
+ CGNL & $ 75.75^{\textcolor{red}{\uparrow}0.42}$ & $ 93.74^{\textcolor{green}{\downarrow}0.23}$ &
+ CGNL & $ 75.27^{\textcolor{green}{\downarrow}0.06}$ & $ 93.05^{\textcolor{green}{\downarrow}0.92}$ \\

\hline
\textbf{+ Ours} & $ \mathbf{\underline{76.34}^{\textcolor{red}{\uparrow}1.01}}$ & $ \mathbf{\underline{94.48}^{\textcolor{red}{\uparrow}0.51}}$ & 
\textbf{+ Ours} & $ \mathbf{\underline{76.41}^{\textcolor{red}{\uparrow}1.08}}$ & $ \mathbf{\underline{94.38}^{\textcolor{red}{\uparrow}0.41}}$ &
\textbf{+ Ours} & $ \mathbf{\underline{76.02}^{\textcolor{red}{\uparrow}0.69}}$ & $ \mathbf{\underline{94.08}^{\textcolor{red}{\uparrow}0.11}}$ \\
\bottomrule
\end{tabular}
\end{table*}
\begin{table*}[!htbp]
\centering
\small
\renewcommand\tabcolsep{3.0pt}
\caption{The Performance of Nonlocal-based Blocks Inserted into Different Position of Deep Networks on CIFAR-100}
\label{Tab:stage}
\begin{tabular}{lcc|lcc|lcc}
\toprule
\multicolumn{3}{c|}{Stage 1} & \multicolumn{3}{c|}{Stage 2} & \multicolumn{3}{c}{Stage 3}\\
\hline
Models & Top-1 (\%) & Top-5 (\%) & Models & Top-1 (\%) & Top-5 (\%)  & Models & Top-1 (\%) & Top-5 (\%) \\
\hline

rereResNet56 & $ 75.33^{\textcolor{red}{\uparrow}0.00}$ & $ 93.97^{\textcolor{red}{\uparrow}0.00}$ & 
PreResNet56 & $ 75.33^{\textcolor{red}{\uparrow}0.00}$ & $ 93.97^{\textcolor{red}{\uparrow}0.00}$ &
PreResNet56 & $ 75.33^{\textcolor{red}{\uparrow}0.00}$ & $ \mathbf{\underline{93.97}^{\textcolor{red}{\uparrow}0.00}}$ \\

+ NL & $ 75.31^{\textcolor{green}{\downarrow}0.02}$ & $ 92.84^{\textcolor{green}{\downarrow}1.13}$ &
+ NL & $ 75.64^{\textcolor{red}{\uparrow}0.31}$ & $ 93.79^{\textcolor{green}{\downarrow}0.18}$ & 
+ NL & $ 75.28^{\textcolor{green}{\downarrow}0.05}$ & $ 93.93^{\textcolor{green}{\downarrow}0.04}$ \\

+ NS & $ 75.83^{\textcolor{red}{\uparrow}0.50}$ & $ 93.87^{\textcolor{green}{\downarrow}0.10}$ &
+ NS & $ 75.74^{\textcolor{red}{\uparrow}0.41}$ & $ 94.02^{\textcolor{red}{\uparrow}0.05}$ &
+ NS & $ 75.44^{\textcolor{red}{\uparrow}0.11}$ & $ 93.86^{\textcolor{green}{\downarrow}0.11}$ \\

+ $\mathrm{A^2}$  & $ 75.58^{\textcolor{red}{\uparrow}0.25}$ & $ 94.27^{\textcolor{red}{\uparrow}0.30}$ &
+ $\mathrm{A^2}$  & $ 75.60^{\textcolor{red}{\uparrow}0.27}$ & $ 93.82^{\textcolor{green}{\downarrow}0.15}$ &
+ $\mathrm{A^2}$   & $ 75.21^{\textcolor{green}{\downarrow}0.12}$ & $ 93.65^{\textcolor{green}{\downarrow}0.32}$ \\

+ CGNL & $ 75.75^{\textcolor{red}{\uparrow}0.42}$ & $ 93.74^{\textcolor{green}{\downarrow}0.23}$ &
+ CGNL & $ 74.54^{\textcolor{green}{\downarrow}0.79}$ & $ 92.65^{\textcolor{green}{\downarrow}1.32}$ &
+ CGNL & $ 74.90^{\textcolor{green}{\downarrow}0.43}$ & $ 92.46^{\textcolor{green}{\downarrow}1.51}$ \\

\hline
\textbf{+ Ours} & $ \mathbf{\underline{76.41}^{\textcolor{red}{\uparrow}1.08}}$ & $ \mathbf{\underline{94.38}^{\textcolor{red}{\uparrow}0.41}}$ &
\textbf{+ Ours} & $ \mathbf{\underline{76.29}^{\textcolor{red}{\uparrow}0.96}}$ & $ \mathbf{\underline{94.27}^{\textcolor{red}{\uparrow}0.30}}$ &
\textbf{+ Ours} & $ \mathbf{\underline{75.68}^{\textcolor{red}{\uparrow}0.35}}$ & $ 93.90^{\textcolor{green}{\downarrow}0.07}$ \\
\bottomrule
\end{tabular}
\end{table*}

\section{Experiments}

In this section, we validate the robustness of nonlocal-based blocks with varying numbers, channels, and positions. Then, we show the performance of the proposed SNL in image classification tasks on Cifar-10/100, ImageNet, action recognition tasks on UCF-101 dataset, and semantic segmentation on Cityscapes dataset. The experimental results of fine-gaining classification on CUB-200 datasets and person re-identification task on ILID-SVID~\cite{ilid_reid}, Mars~\cite{mars_reid}, and Prid-2011~\cite{prid_reid} datasets are given in the appendix 3. All the methods are implemented using PyTorch~\cite{pytorch} toolbox with an Intel Core i9 CPU and $8$ Nvidia RTX 2080 Ti GPUs.


\subsection{Ablation Studies}
\label{sec_4_1}
\noindent \textbf{Experimental Setup.}
Following Tao \etal~\cite{ns}, the robustness testings are conducted on CIFAR-100 dataset containing $50$k training images and $10$k test images of $100$ classes. PreResNet56~\cite{preresnet} is used as the backbone network. Unless otherwise specified, we set $C_s = C_1 / 2$ and add $1$ nonlocal-based block right after the second residual block in the early stage (stage 1). The SGD optimizer is used with the weight decay $10^{-4}$ and momentum  $0.9$. The initial learning rate is $0.1$, which is divided by $10$ at $150$ and $250$ epochs. All the models are trained for $300$ epochs.

\noindent \textbf{The number of channels in transferred feature space.}
The nonlocal-based block firstly reduces the channels of original feature map $C_{1}$ into the transferred feature space $C_{s}$ to reduce computation. The larger $C_{s}$ is, the more redundant information tends to be contained. This introduces the noise when calculating the affinity matrix $\mA$. However, if $C_{s}$ is too small, the output feature map is hard to be reconstructed due to inadequate features. To test the robustness for the value of the $C_{s}$, we generate three types of models with different $C_{s}$ settings: \emph{Non-Reduction} ($C_{s} = C_{1}$), \emph{Half-Reduction} ($C_{s} = C_{1}/2$), and \emph{Quarter Reduction} ($C_{s} = C_{1} / 4$). Table~\ref{Tab:subchannel} shows the experimental results of the $3$ types of models with different nonlocal-based blocks. Our SNL block outperforms other models profited by the flexibility for learning. 

Moreover, from Table~\ref{Tab:subchannel}, we can see that the performance of the CGNL steeply drops when adopts large transferred channels. The reason is that the CGNL block concerns the relations among channels. When the number of the transferred channels increases, the relations among the redundant channels seriously interfere with its effects. Overall, our SNL block is the most robust for a large number of transferred channels (our model rises $1.01\%$ in Top-1 while the best of others only rise $0.18\%$ over the backbone).

\noindent \textbf{The stage/position for adding the nonlocal-based blocks.}
The nonlocal-based blocks can be added into the different stages of the PreResNet to form the Nonlocal Network. Tao \etal~\cite{ns} add them into the early stage of the PreResNet to catch the long-range relations. Here we show the performance of adding different types of nonlocal-based blocks into the $3$ stages (the first, the second, and the third stage of the PreResNet) in Table~\ref{Tab:stage}. We can see that the results of the NL block are lower than the backbones when added into the early stage. However, our proposed SNL block has an averagely $1.08\%$ improvement over the backbone when being added into the early stage, which is more than two times over other types of blocks ($0.42\%$ for the best case). 



\begin{table*}[!htbp]
\centering
\small
\caption{Experiments for Adding Different Types of Nonocal-based Blocks into PreResnet56 and ResNet50 on CIFAR-10/100}
\label{tab:cifar}
\begin{tabular}{lcc|lcc|lcc}
\toprule
 \multicolumn{3}{c}{CIFAR-10} & \multicolumn{6}{|c}{CIFAR-100}\\
\hline
Models & Top-1 (\%) & Top-5 (\%) & Models & Top-1 (\%) & Top-5 (\%)  & Models & Top-1 (\%) & Top-5 (\%) \\
\hline
ResNet50 & $94.94^{\textcolor{red}{\uparrow}0.00}$ & $99.87^{\textcolor{red}{\uparrow}0.00}$ &
PreResnet56 & $75.33^{\textcolor{red}{\uparrow}0.00}$ & $93.97^{\textcolor{red}{\uparrow}0.00}$ & 
ResNet50 & $76.50^{\textcolor{red}{\uparrow}0.00}$ & $93.14^{\textcolor{red}{\uparrow}0.00}$\\

+ NL & $94.01^{\textcolor{green}{\downarrow}0.93}$ & $99.82^{\textcolor{green}{\downarrow}0.05}$ &
+ NL & $75.31^{\textcolor{green}{\downarrow}0.02}$ & $92.84^{\textcolor{green}{\downarrow}1.33}$ &
+ NL & $76.77^{\textcolor{red}{\uparrow}0.27}$ & $93.55^{\textcolor{red}{\uparrow}0.41}$ \\

+ NS & $95.15^{\textcolor{red}{\uparrow}0.21}$ & $99.88^{\textcolor{red}{\uparrow}0.01}$ & 
+ NS & $75.83^{\textcolor{red}{\uparrow}0.50}$ & $93.87^{\textcolor{green}{\downarrow}0.10}$ & 
+ NS & $77.90^{\textcolor{red}{\uparrow}1.40}$ & $\mathbf{\underline{94.34}^{\textcolor{red}{\uparrow}1.20}}$ \\

+ $\mathrm{A^2}$   & $94.41^{\textcolor{green}{\downarrow}0.53}$ & $99.83^{\textcolor{green}{\downarrow}0.05}$ &
+ $\mathrm{A^2}$  & $75.58^{\textcolor{red}{\uparrow}0.25}$ & $94.27^{\textcolor{red}{\uparrow}0.30}$ & 
+ $\mathrm{A^2}$   & $77.30^{\textcolor{red}{\uparrow}0.80}$ & $93.40^{\textcolor{red}{\uparrow}0.26}$ \\

+ CGNL & $94.49^{\textcolor{green}{\downarrow}0.45}$ & $99.92^{\textcolor{red}{\uparrow}0.05}$ &
+ CGNL & $75.75^{\textcolor{red}{\uparrow}0.42}$ & $93.74^{\textcolor{green}{\downarrow}0.23}$ &
+ CGNL & $74.88^{\textcolor{green}{\downarrow}1.62}$ & $92.56^{\textcolor{green}{\downarrow}0.58}$\\
\hline
\textbf{+ Ours} & $\mathbf{\underline{95.32}^{\textcolor{red}{\uparrow}0.38}}$ & $\mathbf{\underline{99.94}^{\textcolor{red}{\uparrow}0.07}}$ &
\textbf{+ Ours} & $\mathbf{\underline{76.41}^{\textcolor{red}{\uparrow}1.08}}$ & $\mathbf{\underline{94.38}^{\textcolor{red}{\uparrow}0.39}}$ &
\textbf{+ Ours} & $\mathbf{\underline{78.17}^{\textcolor{red}{\uparrow}1.67}}$ & $94.17^{\textcolor{red}{\uparrow}1.03}$\\
\bottomrule
\end{tabular}
\end{table*}
\noindent \textbf{The number of the nonlocal-based blocks.} We test the robustness for adding different numbers of the nonlocal-based blocks into the backbone. The results are shown in Table~\ref{tab:number}. ``$\times 3$" means three blocks are added into stage $1,2$, and $3$, respectively, and the accuracy in the brackets represents their results. We can see that adding three proposed SNL operators into different stages of the backbone generates a larger improvement ($1.37\%$) than the NS operator and NL operator. This is because these two models cannot well aggregate the low-level features and interfere with the following blocks when adding NS and NL into the early stage. 
\begin{table}[!htbp]
\centering
\small
\caption{Experiments for Adding Different Number of Nonlocal-based Blocks into PreResNet56 on CIFAR-100}
\renewcommand\tabcolsep{1.7pt}
\label{tab:number}
\begin{tabular}{lc|c}
\toprule
Models & Top-1 (\%) & Top-5 (\%)\\
\hline
PreResNet56 & $ 75.33^{\textcolor{red}{\uparrow}0.00}$ & $ 93.97^{\textcolor{red}{\uparrow}0.00}$\\
+ NL ($\times 3$)& $ 75.31^{\textcolor{green}{\downarrow}0.02}$ ($74.34^{\textcolor{green}{\downarrow}0.99}$) & $ 92.84^{\textcolor{green}{\downarrow}1.13}$ ($ 93.11^{\textcolor{green}{\downarrow}0.86}$)\\

+ NS ($\times 3$)& $ 75.83^{\textcolor{red}{\uparrow}0.50}$ ($75.00^{\textcolor{green}{\downarrow}0.33}$) & $ 93.87^{\textcolor{green}{\downarrow}0.10}$ ($ 93.57^{\textcolor{green}{\downarrow}0.40}$)\\

+ $\mathrm{A^2}$ ($\times 3$) & $ 75.58^{\textcolor{red}{\uparrow}0.25}$ ($ 75.63^{\textcolor{red}{\uparrow}0.33}$) & $ 94.27^{\textcolor{red}{\uparrow}0.30}$ ($\mathbf{\underline{94.12}^{\textcolor{red}{\uparrow}0.15}}$) \\

+ CGNL  ($\times 3$) & $ 75.75^{\textcolor{red}{\uparrow}0.42}$ ( $75.96^{\textcolor{red}{\uparrow}0.63}$) & $ 93.74^{\textcolor{green}{\downarrow}0.23}$ ($ 93.10^{\textcolor{green}{\downarrow}0.87}$)\\

\hline
\textbf{+ Ours}  ($\times 3$) & $\mathbf{\underline{76.41}^{\textcolor{red}{\uparrow}1.08}}$  ($ \mathbf{\underline{76.70}^{\textcolor{red}{\uparrow}1.37}}$)& 
$\mathbf{\underline{94.38}^{\textcolor{red}{\uparrow}0.41}}$ ($93.94^{\textcolor{green}{\downarrow}0.03}$) \\
\bottomrule
\end{tabular}
\end{table}



\subsection{Main Results}
\label{sec_4_2}

\noindent \textbf{Image Classification.}
We use the ResNet50~\cite{resnet} as the backbone and insert the SNL block right before the last residual block of \textit{res4} for a fair comparison. Other settings for the CIFAR-10/100 are the same as the setting discussed in Sec.~\ref{sec_4_1}. For the ImageNet, the SGD optimizer is used with the weight decay $10^{-4}$ and momentum $0.9$. The initial learning rate is set to be $0.01$, which is divided by $10$ at $31$, $61$, and $81$ epochs. All the models are trained for $110$ epoch.
 
Table~\ref{tab:cifar} shows the results on the CIFAR datasets. When adding SNL, it improves the Top-1 accuracy by $0.38\%$ absolutely, which is nearly two times over other nonlocal-based blocks (the best is $0.21\%$). For CIFAR100, using SNL brings significant improvements about $1.67\%$ with ResNet50. While using a more simple backbone PreResnet56, our model can still generate $1.08\%$ improvement which is not marginal compared with previous works~\cite{nl, ns, cgnl}. 

The results of ImageNet are shown in Table~\ref{tab:imagenet}. 
Note that we exhibit other results from their original paper. Our SNL achieves a clear-cut improvement ($1.96\%$) with a minor increment in complexity ($12\%$ and $10\%$ higher in FLOPs and Size respectively) compared with the nonlocal-based blocks. Moreover, our SNL is also superior to other types of blocks such as SE~\cite{senet}, CGD~\cite{cgdnet} and GE~\cite{genet} ($0.11$\% higher in Top-1 and $2.02$M lower in size than the GE block).

\begin{table}
\centering
\small
\caption{Results on ImageNet Dataset}
\label{tab:imagenet}
\begin{tabular}{lccc}
\toprule
Models & Top-1 (\%) & FLOPs (G) & Size (M)\\
\hline
ResNet50 & $76.15^{\textcolor{red}{\uparrow}0.00}$ & $4.14$ & $25.56$ \\ 
+ CGD & $76.90^{\textcolor{red}{\uparrow}0.75}$ & $+ 0.01$ & $+ 0.02$ \\ 
+ SE & $77.72^{\textcolor{red}{\uparrow}1.57}$ & $+ 0.10$ & $+ 2.62$\\
+ GE & $78.00^{\textcolor{red}{\uparrow}1.85}$ & $+0.10$ & $+5.64$ \\
\cline{2-4}
+ NL & $76.70^{\textcolor{red}{\uparrow}0.55}$ & $+ 0.41$ & $+ 2.09$\\
+ $\mathrm{A^2}$ & $77.00^{\textcolor{red}{\uparrow}0.85}$  & $+ 0.41$ & $+ 2.62$\\
+ simpNL\cite{gcnet} & $77.28^{\textcolor{red}{\uparrow}1.13}$ & - & $+ 1.05$ \\
+ CGNL & $77.32^{\textcolor{red}{\uparrow}1.17}$ & $+ 0.41$ & $+ 2.09$\\
\hline
 \textbf{+ Ours} & $\mathbf{\underline{78.11}^{\textcolor{red}{\uparrow}1.96}}$ & $+0.51$ & $+ 2.62$\\
\bottomrule
\end{tabular}
\end{table}

\begin{figure}
    \centering
    \includegraphics[width=0.48\textwidth]{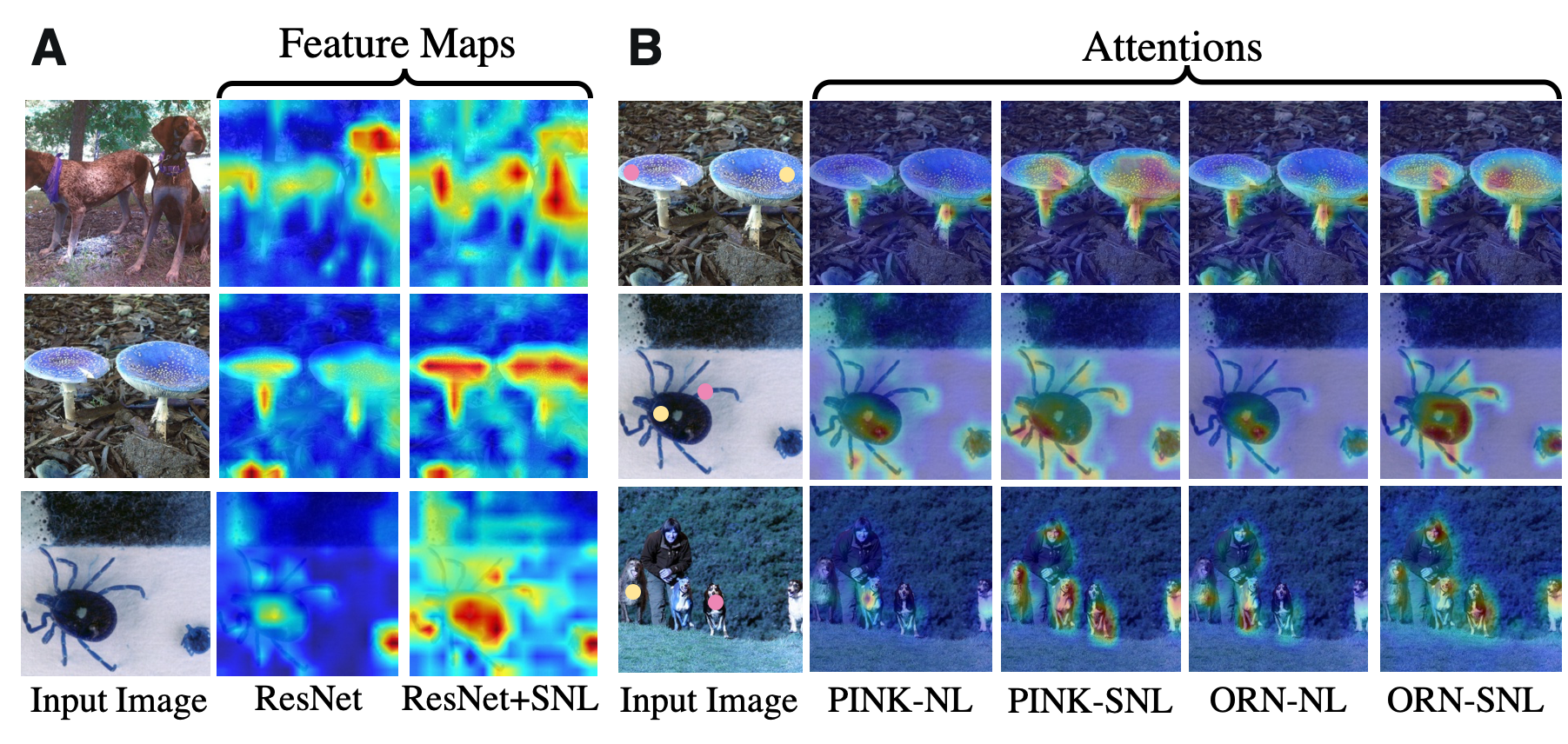}
    \caption{
    \textbf{ \sf{A}.} The visualization of the feature maps when adding SNL into the the backbone.
    \textbf{ \sf{B}.} The visualization of the attention maps for two positions (``Pink" and ``Orange" dots). The heatmaps show the strength of similarity.
    }
    \label{fig:imagenet_feature_attention}
\end{figure}

We also visualize the output feature maps of the ResNet50 with SNL and the original ResNet50 in Fig.~\ref{fig:imagenet_feature_attention} \textcolor{red}{\textbf{\sf{A}}}. Benefited from the rich and structured information considered in SNL, the response of the similar features among long-range spatial positions are enhanced as shown in the two mushrooms, balls, and those animals. Moreover, Fig.~\ref{fig:imagenet_feature_attention} \textcolor{red}{\textbf{\sf{B}}} shows the attention maps produced by our SNL and the original NL block where the ``Pink" and ``Orange" dots are the central positions and the heatmaps represent the similarity between the central position and other positions. Compared with the original NL block, SNL can pay more attention to the crucial parts than the original NL block profited by the better approximation formation as discussed in Sec.~\ref{sec_3_3}.

\noindent \textbf{Action Recognition.}
Experiments are conducted on the UCF-101 dataset, which contains $7912$ training videos and $1625$ test videos of $101$ different human actions. Our SNL block is tested on the UCF-101 dataset for capturing the dependence for the temporal frames. We follow the I3D structure~\cite{I3d} which uses $k \times k \times k$ kernels to replace the convolution operator in the residual block to learn seamless spatial-temporal feature extractors. The weights are initialized by the pre-trained I3D model on Kinetics dataset~\cite{kinetics}. Inserting nonlocal-based blocks into the I3D can help to capture the relations among frame pairs with long distance and improve the feature representation. In the training process, each input video clip is cropped into $124*124*16$ to train our model. An initial learning rate of $0.01$ is adopted, which is subsequently divided by $10$ each $40$ epoch. The training stops at the $100$ epoch. Other hyper-parameters of the experimental setup are the same in Sec.~\ref{sec_4_1}. 

\begin{table}[!htbp]
\centering
\small
\caption{Results on UCF-101 Datasets}
\label{tab:cub_ucf}
\renewcommand\tabcolsep{1.7pt}
\begin{tabular}{ccccc}
\toprule
Models & Top-1 (\%) & Top-5 (\%) & FLOPs (G) & Size (M) \\
\hline
I3D  & $81.57^{\textcolor{red}{\uparrow}0.00}$ & $95.40^{\textcolor{red}{\uparrow}0.00}$ & 10.10 & 47.02\\
+ NL &  $82.88^{\textcolor{red}{\uparrow}1.31}$ & $95.74^{\textcolor{red}{\uparrow}0.34}$  & + 0.21 & + 2.10 \\
+ NS  & $82.50^{\textcolor{red}{\uparrow}0.93}$ &  $95.84^{\textcolor{red}{\uparrow}0.44}$  & + 0.26  & + 2.10\\
+ $\mathrm{A^2}$  & $82.68^{\textcolor{red}{\uparrow}1.11}$ &  $95.85^{\textcolor{red}{\uparrow}0.45}$  & + 0.21 & + 2.10\\
+ CGNL  & $83.38^{\textcolor{red}{\uparrow}1.81}$ & $95.42^{\textcolor{red}{\uparrow}0.02}$  & + 0.21 & + 2.10\\
\hline
\textbf{+ Ours} & $\mathbf{\underline{84.39}^{\textcolor{red}{\uparrow}2.82}}$ & $\mathbf{\underline{97.66}^{\textcolor{red}{\uparrow}2.26}}$   & + 0.26 & + 2.62 \\
\bottomrule
\end{tabular}
\end{table}

Table~\ref{tab:cub_ucf} (UCF-101) shows the clip-level Top1 and Top5 metrics on the action recognition. The network with our SNL generates significant improvements ($2.82\%$) than the backbone and outperforms all other nonlocal-based models. It shows that our proposed SNL is also effective to catch the long-range dependencies among the temporal frames. We also conduct the experiments on UCF-101 dataset with other state-of-the-art action recognition models in appendix 3.

\noindent \textbf{Semantic Segmentation.} NL-based blocks are widely used in semantic segmentation due to the requirement of large respective fields. Thus, both experiments and ablation studies are conducted on Cityscapes dataset, which contains 2975 train images and 500 validation images of the urban scene. For a fair comparison, the ``mmsegmentation toolbox"~\cite{mmseg} is used with the recommendation hyper-parameter for NL-based blocks. Specifically, the ``ResNet50 with two FCN decoders"~\cite{fcn} were used as the baseline model, where the first FCN was replaced by NL-based blocks with setting $C_s = C_1 / 4$. SGD optimizer with momentum $0.9$, weight decay $5$e$-4$, and initialization learning rate $0.01$ are used to train all the NL-based networks for $40$k iterators, where poly strategy with power $0.9$ is used to adjust the learning rate.

Table~\ref{tab:cityscape} shows the experimental results of NL-based networks on Cityscapes dataset. To evaluate the effectiveness of the two modifications of the proposed SNL, two new blocks SNL$_{a1}$ and SNL$_{a2}$ are conducted to consider the symmetrization affinity or ``the 1$^{st}$-order" Chebyshev term respectively. It can be seen that compared with the original NL block, our (SNL$_{a1}$) improves 1.61 in mIoU and 1.75 in mAcc with no complexity increasing benefited from confirming the existence of the graph spectral domain. As for the effect of Chebyshev term,  using ``the 1$^{st}$-order " (SNL$_{a2}$) is the best, and the ``sharing weight 1$^{st}$-order" (NS) is better than using ``only the 1$^{st}$-order term" (NL). Besides, when combining these two factors, the performance of our SNL is 0.68 higher in mIoU and 0.14 higher in mAcc with only 0.13M higher model size than the state-of-the-art DNL block~\cite{DNL}.

\begin{table}
\renewcommand\tabcolsep{1.0pt} 
\centering
\small
\caption{Experiments for Nonocal-based Blocks Added into ResNet50-FCN on Cityscapes Dataset}
\label{tab:cityscape}
\begin{tabular}{lcccc}
\toprule
Models & mIoU(\%) & mAcc(\%) & FLOPs(G) & Size(M)\\
\hline
Backbone & $69.19^{\textcolor{red}{\uparrow}0.00}$ & $76.60^{\textcolor{red}{\uparrow}0.00}$ & $395.76 $ & $47.13$\\
\hline
NL & $74.15^{\textcolor{red}{\uparrow}4.96}$ & $81.83^{\textcolor{red}{\uparrow}5.23}$ & $+4.31$ & $+0.53$\\
NS & $75.44^{\textcolor{red}{\uparrow}6.25}$ & $83.36^{\textcolor{red}{\uparrow}6.76}$ & $+5.37$ & $+0.53$\\
CC & $75.34^{\textcolor{red}{\uparrow}6.15}$ & $82.75^{\textcolor{red}{\uparrow}6.15}$ & $+5.37$ & $+0.33$\\
DNL & $76.19^{\textcolor{red}{\uparrow}7.00}$ & $84.61^{\textcolor{red}{\uparrow}8.01}$ & $+4.31$ & $+0.53$\\
 \hline
\textbf{SNL$_{a_1}$} & $75.76^{\textcolor{red}{\uparrow}6.57}$ & $83.58^{\textcolor{red}{\uparrow}6.98}$ & $+4.31$ & $+0.53$\\
\textbf{SNL$_{a_2}$} & $75.94^{\textcolor{red}{\uparrow}6.75}$ & $83.85^{\textcolor{red}{\uparrow}7.25}$ & $+5.37$ & $+0.66$\\
 \hline
\textbf{SNL} & \bm{$\underline{76.87}^{\textcolor{red}{\uparrow}7.68}$} &  \bm{$\underline{84.75}^{\textcolor{red}{\uparrow}8.15}$} & $+5.37$  & $+0.66$\\
\bottomrule
\end{tabular}
\end{table}


\section{Conclusion}
This paper provides a novel perspective for the model design of nonlocal-based blocks. In this context, we make the connection between different variants of nonlocal methods and graph signal processing. Five well-known nonlocal blocks are unified and interpreted under the perspective of graph filters. A novel nonlocal block called SNL with stronger theoretical basis is proposed. It works universally well across a wide array of vision tasks and offers better performance than other nonlocal-based counterparts. Future work will focus on designing novel nonlocal blocks based on our spectral view and extending our spectral view on other self-attention based modules such as the vision transformer.  

\section{Acknowledgements} This work was supported by the Shenzhen Science and Technology Program (1210318663); the National Biomedical Imaging Facility Grant; the Shenzhen Nanshan Innovation and Business Development Grant; the NSFC Grants (62072449, 61632003)

{\small
\bibliographystyle{ieee_fullname}
\bibliography{egbib}
}

\end{document}